%% file: main.tex
\documentclass[runningheads]{llncs}

\usepackage[mobile]{eccv}

\usepackage{graphicx}
\usepackage{booktabs}

\usepackage[accsupp]{axessibility}  % Improves PDF readability for those with disabilities.

\usepackage{multirow}
\usepackage{adjustbox}
\usepackage{enumitem}
\usepackage{wrapfig}
\usepackage{soul}
\usepackage{siunitx}
\usepackage{xcolor}
\usepackage{colortbl, array}
\usepackage{pgfplotstable}
\pgfplotsset{compat=1.8}

\definecolor{rulecolor}{RGB}{0,71,171}
\definecolor{tableheadcolor}{gray}{0.92}
\newcommand{\topline}{ %
        \arrayrulecolor{rulecolor}\specialrule{0.1em}{\abovetopsep}{0pt}%
        \arrayrulecolor{tableheadcolor}\specialrule{\belowrulesep}{0pt}{0pt}%
        \arrayrulecolor{rulecolor}}
\newcommand{\midtopline}{ %
        \arrayrulecolor{tableheadcolor}\specialrule{\aboverulesep}{0pt}{0pt}%
        \arrayrulecolor{rulecolor}\specialrule{\lightrulewidth}{0pt}{0pt}%
        \arrayrulecolor{white}\specialrule{\belowrulesep}{0pt}{0pt}%
        \arrayrulecolor{rulecolor}}
\newcommand{\bottomline}{ %
        \arrayrulecolor{white}\specialrule{\aboverulesep}{0pt}{0pt}%
        \arrayrulecolor{rulecolor} %
        \specialrule{\heavyrulewidth}{0pt}{\belowbottomsep}}%

\pgfplotstableset{normal/.style ={%
        header=true,
        string type,
        column type=l,
        every odd row/.style={
            before row=
        },
        every head row/.style={
            before row={\topline\rowcolor{tableheadcolor}},
            after row={\midtopline}
        },
        every last row/.style={
            after row=\bottomline
        },
        col sep=&,
        row sep=\\
    }
}
\usepackage[pagebackref,breaklinks,colorlinks]{hyperref}
\usepackage{orcidlink}

\begin{document}

\title{LORD: Large Models based Opposite Reward Design for Autonomous Driving} 

\titlerunning{LORD:Large Models based Opposite Reward
Design}

\author{Xin Ye$^*$ \
Feng Tao\thanks{Equal contributions.$^\dagger$ Corresponding author} \
Abhirup Mallik \ Burhaneddin Yaman$^\dagger$ \ Liu Ren}

\authorrunning{ X. Ye et al.}

\institute{Bosch Research North America \& \\
Bosch Center for Artificial Intelligence (BCAI) \\ 
\email{\{xin.ye3, feng.tao2, abhirup.mallik, burhaneddin.yaman, liu.ren\}@us.bosch.com}
}

\maketitle

\input{eccv2024/Sections/abstract}

\input{eccv2024/Sections/introduction}

\input{eccv2024/Sections/related_work}
\input{eccv2024/Sections/approach}

\input{eccv2024/Sections/experiments}

\input{eccv2024/Sections/conclusion}

\bibliographystyle{splncs04}
\bibliography{ref}

\clearpage
\input{eccv2024/Sections/appendix}

\end{document}

%% file: eccv2024/Sections/abstract.tex
\begin{abstract}
Reinforcement learning (RL) based autonomous driving has emerged as a promising alternative to data-driven imitation learning approaches. However, crafting effective reward functions for RL poses challenges due to the complexity of defining and quantifying good driving behaviors across diverse scenarios. Recently, large pretrained models have gained significant attention as zero-shot reward models for tasks specified with desired linguistic goals.
However, the desired linguistic goals for autonomous driving such as ``\textit{drive safely}'' are ambiguous and incomprehensible by pretrained models. On the other hand, undesired linguistic goals like ``\textit{collision}'' are more concrete and tractable. In this work, we introduce LORD, a novel large models based opposite reward design through undesired linguistic goals to enable the efficient use of large pretrained models as zero-shot reward models.  Through extensive experiments, our proposed framework shows its efficiency in leveraging the power of large pretrained models for achieving safe and enhanced autonomous driving.  Moreover, the proposed approach shows improved generalization capabilities as it outperforms counterpart methods across diverse and challenging driving scenarios. 
\end{abstract}

%% file: eccv2024/Sections/introduction.tex
\section{Introduction}\label{sec:intro}

Autonomous driving is a challenging task that demands both deep comprehension of the environment and ability to swiftly reacting to changes. Rapid advancements in deep learning have triggered significant progress in this domain, mainly through imitation learning (IL) approaches \cite{chen2020learning, hu2023_uniad,jiang2023vad,pan2024vlp}. Despite showing impressive results, the performance of these IL methodologies heavily relies on the size of data \cite{quinonero2008dataset, codevilla2019exploring}. Thus, IL approaches are inherently subject to dataset bias and lack rational in decision making. To address these challenges, reinforcement learning (RL) based approaches that optimize driving policies by interacting with the environment and maximizing the rewards have gathered growing interest as alternatives for autonomous driving tasks \cite{sallab2017deep, kendall2019learning, kiran2021deep}.

Reinforcement learning thrives when paired with effective reward functions, which serve as the guiding principles for learning optimal behaviors\cite{ziebart2008maximum,gail,vice,christiano2017deep,fu2017learning,hejna2023few}. However, crafting these reward functions often proves costly, particularly when relying on human feedback for their formulation \cite{christiano2017deep, zhan2021human}. Additionally, manually specifying such reward functions presents a formidable challenge in avoiding reward hacking\cite{pan2021effects}. This challenge is further compounded in the autonomous driving tasks due to the difficulty of defining and quantifying good driving behavior, as well as generalizing them across diverse driving scenarios. To address these challenges, leveraging large pretrained models emerges as a promising solution for crafting efficient and generalizable reward functions for autonomous driving systems.

Large pretrained models exhibit human-like reasoning abilities and have demonstrated remarkable performance across a spectrum of tasks \cite{radford2021learning,florence,flamingo,Palme,rt2,min2023recent,yang2023foundation}. In the realm of robotics, the integration of these models in reward functions have shown good performance and promising generalization capabilities \cite{vlmrm,roboclip,ellm}. In these works, experiments evolve around tasks where the desired goal state is either known or can be easily defined.  Thus, describing the desired goal state in form of a linguistic goal which is comprehensible by pretrained models enables exploiting the pretrained models as zero-shot reward models.  While these approaches show good performance in variety of robotic tasks, they encounter significant challenges in autonomous driving. In such intricate scenarios, direct linguistic goals become particularly arduous for large models to grasp, highlighting the need for more nuanced strategies to ensure effective comprehension and decision-making.

In this work, we present a novel approach to reward design for safe and enhanced autonomous driving: the concept of opposite reward design through undesired linguistic goals in order to leverage large pretrained models as zero-shot reward models. In autonomous driving scenarios, linguistically defining desired goal state such as ``\textit{drive safely}'' can be ambiguous and challenging. However, undesired linguistic goals, such as ``\textit{collision}'', offer a more tangible and understandable objective for both humans and large pretrained models. By introducing opposite reward design, we aim to enhance the interpretability, generalizability and effectiveness of autonomous driving systems, making them more capable of navigating complex environments while prioritizing safety. To harness the full potential of our approach, we construct a closed-loop driving environment. We conduct extensive experiments on large pretrained image, video, and language models to evaluate the efficacy of our proposed framework for closed-loop autonomous driving tasks. Notably, our framework achieves significantly improved performance over counterpart methods across various driving scenarios. 

The main contributions of this work are summarized as follows:
\begin{itemize}
    \item We propose LORD, a Large models based Opposite Reward Design, which addresses the ambiguity of linguistic goals in autonomous driving with comprehendable undesired linguistic goals. To the best of our knowledge, this is the first work that leverages large pretrained models with undesired goals in embodied AI domain.

    \item LORD leverages large pretrained image, video and language models with a cosine distance objective for an efficient reward function design for RL based autonomous driving. 

    \item Through extensive experiments, we show LORD consistently achieves significantly improved generalization performance over counterpart methods across various challenging driving scenarios.

\end{itemize}

%% file: eccv2024/Sections/related_work.tex
\section{Related Work}\label{sec:related_work}

\subsection{Reward Design for Reinforcement Learning}
Reward function plays a pivotal role in reinforcement learning, dictating the behavior of autonomous agents. Unlike games where rewards occur naturally, creating a reward function for real-world tasks needs an intentional design that requires extensive expert supervision. The difficulty motivates many researchers to directly learn a reward function by observing a human expert performing the task \cite{ng2000algorithms, abbeel2004apprenticeship,ziebart2008maximum,finn2016guided}. However, these approaches become overly complex when applied to tasks with high dimensional state and action space. More recently, some work leverage discriminator networks with demonstration sets to assign rewards based on the likelihood of a state belonging to the demonstration set \cite{gail,vice,fu2017learning}. Training these discriminator networks still requires a substantial number of expert demonstrations which is not always feasible due to the limited availability of such demonstrations.
Conversely, another line of work involves using human pairwise preferences over data samples to learn the reward function \cite{christiano2017deep,hejna2023few}. While these methods offer good results in some tasks, they often rely on either a large number of valid goal states or significant human effort, making them impractical for many applications, particularly in the context of autonomous driving, where efficiency and scalability are paramount.

\subsection{Reward Design with Large Pretrained Models}
Large pretrained models have recently gained interest as an alternative way for reward design. Describing the goal state through language has been the centerpiece for designing powerful zero-shot reward models. A line of work in this direction has focused on using large language models (LLMs) \cite{kwon2023reward,hu2023language,ellm}. 
For example, ELLM \cite{ellm} utilizes LLMs to reward agent for achieving goals suggested by LLMs. In another work, LLMs have been used as a proxy reward function to capture human preferences by prompting desired behaviors \cite{kwon2023reward}. More recent works use large pretrained multi-modal models. Among these works, VLM-RMs \cite{vlmrm} and RoboCLIP \cite{roboclip} leverages vision language models and video language models, respectively, with desired linguistic goals for the reward design. However, these works have focused on specific robotic applications where the desired goal state exists and is comprehensible by large pretrained models \cite{roboclip,vlmrm}. In contrast, our work focuses on autonomous driving where desired goal states either does not exist or not comprehensible by large pretrained models due to the innate ambiguity of desired linguistic goals.

\subsection{Language Models in Autonomous Driving}
The success of language models in robotics have sparked the interest for incorporation of language models in autonomous driving \cite{pan2024vlp,gptdriver,drivegpt4,drivelm}.  Several works have leveraged LLMs for explainable autonomous driving. LINGO-1 presents a commentator model for reasoning and explainability by training a model combining language with vision and action \cite{lingo1}. Similarly, DriveLM introduces a visual question answering approach to interpret driving actions \cite{drivelm}. Other works have focused on enhancing the planning performance for autonomous driving through language. VLP introduces a plug-in approach by incorporating LLMs with contrastive learning objective into vision-only end-to-end autonomous driving systems \cite{pan2024vlp}.  Several GPT-based driver agents have also been introduced as an alternative to  existing IL and RL based driver agents \cite{drivegpt4,gptdriver,wen2023dilu}. Among these works, DiLu presents a systematic framework which combines reasoning and reflection mechanism to improve the decision making capability of the driver agent \cite{wen2023dilu}. Nevertheless, the substantial reliance on GPT models presents several limitations, such as the occurrence of hallucinations due to the lack of grounding, as well as latency issues that are crucial for real-world deployment. Unlike these approaches, our work utilizes an RL agent with a reward mechanism based on large pretrained models to circumvent these challenges. 

%% file: eccv2024/Sections/approach.tex
\section{Methodology}
\label{sec:approach}
\begin{figure}[ht!]
    \centering
    \includegraphics[width=1\textwidth]{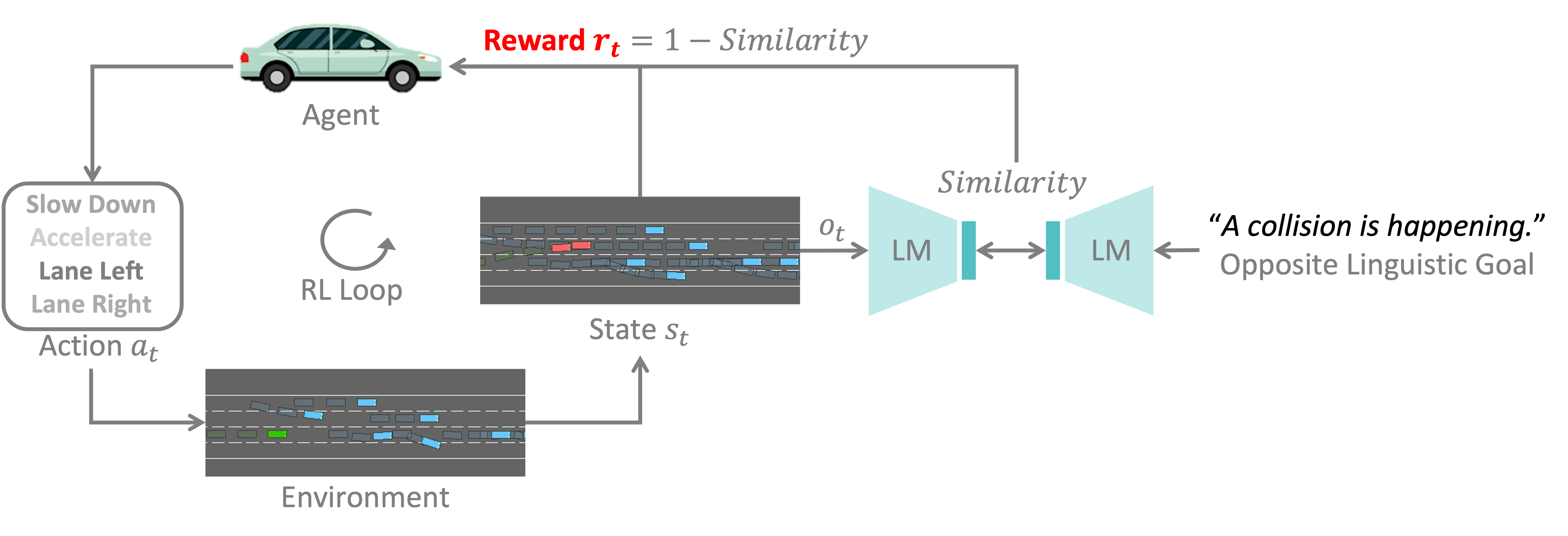}
    \caption{An overview of our LORD powered reinforcement learning framework for closed-loop autonomous driving task. LORD firstly measures cosine similarity between agent's state and undesired goal state using large pretrained models. Followingly, it returns cosine distance as the reward to the agent.}
    \vspace{-0.5cm}
    \label{fig:overview}
\end{figure}

LORD utilizes large pretrained models to generate step-wise rewards for autonomous agents (i.e. ego vehicles) aiming to encourage desired driving behavior and outcomes. This is done by evaluating how different the state of the autonomous agent at each time step is from the undesired goal state described by our opposite linguistic goal for the first time. Moreover, since the state of the autonomous agent can be observed as an image, a video and a linguistic description, we investigate vision-and-language, video-and-language and language models for embedding the agent's state and undesired goal state, respectively. Cosine similarity between the agent's and goal state's embeddings is calculated. Followingly, the agent receives cosine distance (i.e., $1$ - cosine similarity) as the reward for each step. We integrate LORD with reinforcement learning algorithm for closed-loop autonomous driving task. Fig.~\ref{fig:overview} shows an overview of our LORD powered RL framework. Details of our method are presented in the following subsections. We first provide a problem formulation of the closed-loop autonomous driving task in Sec.~\ref{sec:prob}. Subsequently, we describe our proposed LORD and its integration with RL in Sec.~\ref{sec:lord} and Sec.~\ref{sec:rl_lord}, respectively.

\vspace{-0.2cm}

\subsection{Problem Formulation}
\label{sec:prob}
We formulate the closed-loop autonomous driving task as a Partially Observable Markov Decision Process (POMDP) problem defined by a $7$-tuple $<S, O, \theta, A, T, \\R, \gamma>$. Specifically, $S$ is the state space of the ego vehicle.  $O$ is the observation space that is determined by emission function $\theta: S \times O \to [0,1]$. $A$ is a set of actions used to drive the ego vehicle. $T: S\times A \times S \to [0,1]$ is a state transition probability function. $R: \Omega \times A \to \mathbb{R}$ is a reward function to reward desired driving behavior or outcomes, and $\gamma \in (0, 1]$ is a discount factor. To learn a good driving policy $\pi(a_t|s_t)$ informing the ego vehicle which action $a_t$ to take at the state $s_t$, we maximize the expected discounted cumulative rewards $\mathbb{E}[\sum_t^\infty \gamma^t r_{t+1}(a_t,o_{t+1})|s_t]$. The focus of this paper is to efficiently and effectively define the reward function $R$ based on the ego vehicle's observation $O$ and our opposite linguistic goal noted as $goal$. Note that the driving policy $\pi$ is still learned from the ego vehicle's state $S$.

\vspace{-0.2cm}
\subsection{Large Models based Opposite Reward Design}
\label{sec:lord}
\subsubsection{Opposite Linguistic Goal.}
\begin{wrapfigure}{r}{0.45\textwidth}
    \centering
    \vspace{-0.75cm}
    \includegraphics[width=0.35\textwidth]{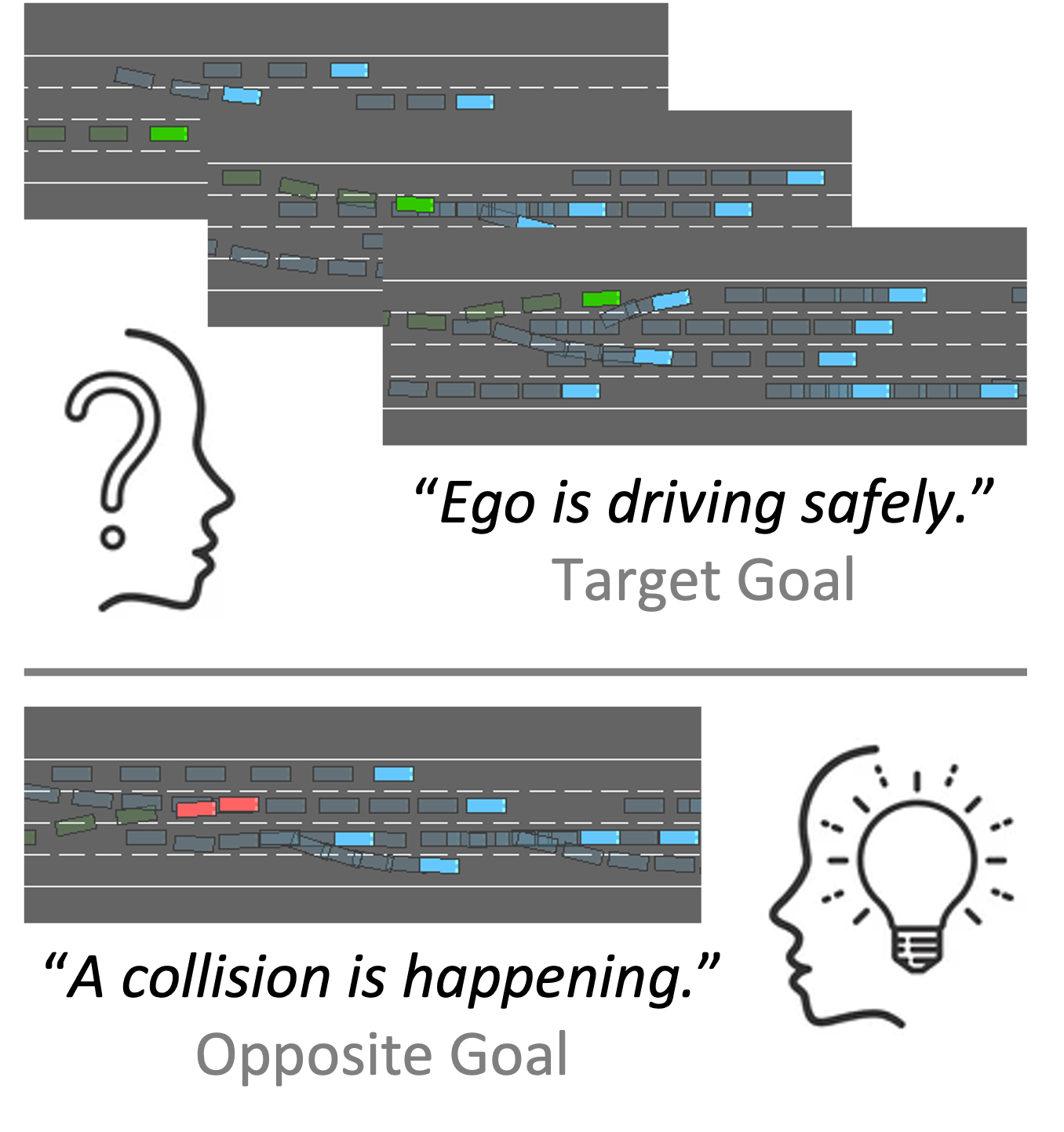}
    \vspace{-0.1cm}
    \caption{The insight of using an opposite goal. In autonomous driving tasks, desired goal states such as ``\textit{drive safely}'' are ambiguous to grasp, whereas undesired goal states such as ``\textit{collision}'' are tractable more comprehensible to humans and large pretrained models.}
    \label{fig:opposite_goal}
    \vspace{-0.4cm}
\end{wrapfigure}
Recent work in the field of robotics has shown a great success in taking large pretrained models as a zero-shot reward model for robotic tasks \cite{vlmrm,roboclip}. An essential element to the success is that they are able to describe the desired task and expected goal state with linguistic descriptions accurately. For example, a task of ``\textit{a humanoid robot kneeling}'' is actually an accurate description to the expected goal state \cite{roboclip}.
In this case, when large pretrained models project the agent's current state and the desired goal state into an embedding space, the distance between the two embeddings forms a natural reward measuring how close the agent is to the desired goal state. However, for autonomous driving task, while the ultimate target is to let the ego vehicle drive safely, it is difficult to describe concretely what the goal state is since there are infinite ways to keep safe in driving. As a result, the embedding of the abstract target goal ``\textit{ego is driving safely}'' is not semantically comparable to the ego vehicle's states and observations. On the contrary, describing unexpected states that the ego vehicle should avoid is more tractable. For example, we can easily imagine what a ``\textit{collision}'' looks like and ground it into an observation. Fig.~\ref{fig:opposite_goal} better illustrates our insight. To this end, we propose to use an opposite linguistic goal ``\textit{a collision is happening}'' as opposed to the target goal ``\textit{ego is driving safely}''.

\subsubsection{Observation Representations}
\label{sec:obs_rep}
The ego vehicle's observation $o_t$ at time step $t$ can be represented in multiple ways: (1) an image capturing the spatial information around the ego vehicle; (2) a video containing both spatial and temporal information with respect to the ego vehicle; (3) a linguistic description to the ego vehicle's current situation.

\begin{itemize}[itemsep=10pt]
    \item \textbf{Image based Observation.} Using an image to represent an observation is straightforward and efficient. A raw image can be obtained directly from camera-like sensors and a more informative image, like Bird's Eye View (BEV) can be built through advanced computer vision method \cite{li2022bevformer}. The high dimensional image based observation contains rich information including surrounding objects and road elements and thus has been wildly adopted for autonomous driving tasks \cite{hu2023_uniad, pan2024vlp}. In this work, without loss of generality, we use the image rendered by the deployment environment as the observation. To project the image based observation and our opposite linguistic goal into the same embedding space, we adopt the vision-and-language model CLIP \cite{radford2021learning} to encode the observation and the goal with its image and text encoders respectively. In particular, we select the CLIP model pretrained on the large-scale dataset LAION-2B \cite{schuhmann2022laion} as it has been shown to have superior performance in \cite{vlmrm}.

    \item \textbf{Video based Observation.} A single static image may not be able to capture the kinematic information, such as the speed and the acceleration of both the ego and npc vehicles. These information is critical for autonomous driving. For example, a vehicle is more likely to collide in a congested traffic scenario if it is driving at a higher speed and with a greater acceleration. To get an observation containing the kinematic information, we generate a video by stacking the latest $30$ consecutive frames of the images at each time step $t$. To compare the video based observation and our opposite linguistic goal, we utilize the video-and-language model S3D \cite{xie2018rethinking}. S3D is pretrained on HowTo100M dataset \cite{miech2019howto100m} which consists of diverse short clips of human demonstrators performing daily tasks. Therefore, the video encoder and text encoder of S3D can encode our video based observation and opposite linguistic goal into a semantically meaningful latent space.

    \item \textbf{Text based Observation.} Inspired by the recent success of large language models (LLMs) and their applications, recent work starts to use textual scenario descriptions as the observations \cite{ellm, wen2023dilu}. In this way, they can leverage LLMs' exceptional human-level abilities to perform the driving task by asking LLMs to make decisions upon the text based observations. In our work, we describe the ego vehicle's current situation in terms of potential collisions. Specifically, we calculate the time to collision (ttc) with each of the surrounding vehicles based on their distance and speed difference. If the ttc is smaller than a predefined threshold, we describe it in our text based observation by ``\textit{A collision will be happening in \{ttc\} seconds.}''. We also describe conditional collisions by ``\textit{A collision would happen in \{ttc\} seconds if ego makes a left/right lane change.}''. To determine how similar the text based observation and our opposite linguistic goal are, we adopt a language model to convert them into embeddings that capture their semantic information. In practice, we use the pretrained SentenceBERT model \cite{reimers2019sentence} which is designed to derive semantically meaningful sentence embeddings that can be compared using cosine-similarity.

\end{itemize}

\subsubsection{Opposite Reward Generation.}
Given our opposite linguistic goal $goal$ and the observation $o_t$ at time step $t$, we define our reward $r_t$ as follows,
\begin{equation}\label{eq:sim}
    similarity(o_t, goal) =  \frac{LM^{o}(o_t) \cdot LM^{g}(goal)}{||LM^{o}(o_t)|| \cdot ||LM^{g}(goal)||}
\end{equation}
\begin{equation}\label{eq:reward}
    r_t = 1 - similarity(o_t, goal)
\end{equation} 
where $LM^{o}$ and $LM^{g}$ denote the large pretrained models used to encode the observation and the opposite linguistic goal, respectively. Our model choices are specified in Sec.~\ref{sec:obs_rep}. Here, we adopt the cosine distance between the observation embedding $LM^{o}(o_t)$ and the opposite linguistic goal embedding $LM^{g}(goal)$ to quantify the reward value for the ego vehicle. In this way, when the ego vehicle is further away from the undesired situation described by our opposite linguistic goal, the ego vehicle can get a higher reward.

\subsection{RL Training with LORD}
\label{sec:rl_lord}
LORD can be integrated with any standard reinforcement learning algorithms. In this work, we follow \cite{xi2022graph}, the latest state-of-the-art reinforcement learning work in autonomous driving domain, by adopting Proximal Policy Optimization (PPO) \cite{schulman2017proximal} algorithm to learn an optimal driving policy $\pi(a_t|s_t)$ for the ego vehicle. LORD is only used in training. During testing, we evaluate the ego vehicle by performing the action $a_t \sim \pi(a_t|s_t)$ at time step $t$ that only depends on the state $s_t$.

%% file: eccv2024/Sections/experiments.tex
\section{Experiments} \label{sec:exp}
We conduct experiments on Highway-env \cite{highway-env} to validate our LORD framework for closed-loop autonomous driving task. In particular, we aim to seek the answers to the following questions:
\begin{itemize}
    \item Is our LORD framework effective in addressing the closed-loop autonomous driving task?
    \item How does each variant of our method work for the closed-loop autonomous driving task?
    \item Does our opposite reward design contribute to the success of our method?
\end{itemize}
We first introduce our experiment setting in Sec.~\ref{sec: exp_setting}, and we rearrange the remaining of this section to answer each of these questions. In Sec.~\ref{sec:baseline_compare}, we compare LORD with baseline methods to answer the first question. We answer the second question in Sec.~\ref{sec:reward_analyze} by conducting an in-depth analysis of the reward values generated by LORD. Ablation studies are provided in Sec.~\ref{sec:reward_ablation} to answer the third question. Sec.~\ref{sec:qual_res} presents qualitative results .

\subsection{Experiment Setting}
\label{sec: exp_setting}
\subsubsection{Simulation Environment.} We adopt Highway-env \cite{highway-env} to conduct all experiments. Highway-env is a well-established simulation platform for closed-loop autonomous driving task in which npc vehicles can react to ego's behavior. It is wildly used in the research work of autonomous driving \cite{wen2023dilu, xi2022graph}. We follow \cite{xi2022graph} to set up the environment. More particularly, we define the state space $S$ of the ego vehicle as its kinematic observation which is a $V \times F$ array provided by the environment that describes a list of $V$ nearby vehicles by a set of features of size $F$, including the vehicles' positions, speeds and orientations. We adopt the discrete meta-actions as the ego vehicle's action space $A$ that consists of lane and speed change. In addition, we also create various traffic situation by setting the density of vehicles and the number of lanes as \cite{wen2023dilu} does to test all methods. Detailed configurations can be found in the supplementary materials.  

\subsubsection{Domain Adaptation.} 
\begin{figure}
\centering
    \begin{subfigure}{0.24\textwidth}
        \includegraphics[width=\textwidth]{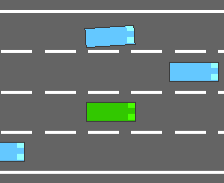}
        \caption{Original driving.}
        \label{fig:od}
    \end{subfigure}
    \begin{subfigure}{0.24\textwidth}
        \includegraphics[width=\textwidth]{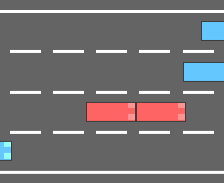}
        \caption{Original colliding.}
        \label{fig:oc}
    \end{subfigure}
    \begin{subfigure}{0.24\textwidth}
        \includegraphics[width=\textwidth]{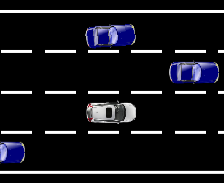}
        \caption{Modified driving.}
        \label{fig:md}
    \end{subfigure}
    \begin{subfigure}{0.24\textwidth}
        \includegraphics[width=\textwidth]{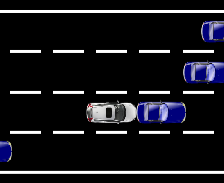}
        \caption{Modified colliding.}
        \label{fig:mc}
    \end{subfigure}
\caption{Illustrations of the original and the modified Highway-env. In the modified environment, white car denotes the ego vehicle and blue cars depict the npc vehicles.}
\label{fig:env}
\end{figure}

Highway-env provides a simplified visualization. All vehicles are depicted as rectangles where the ego vehicle is colored in green and npc vehicles are colored in blue (see Fig.~\ref{fig:od}). When a collision happens, the victim vehicles are colored in red as Fig.~\ref{fig:oc} illustrates. These rendered images are likely out of the training distribution of the large pretrained models. In consequence, our large model based rewards may not work well for the settings with image or video based observations. To remedy this issue, we modify the graphics of the Highway-env by replacing the rectangle textures with more photorealistic car images. Besides, we also remove the useless background of the images. Fig.~\ref{fig:md} and Fig.~\ref{fig:mc} show the snapshots of our modified Highway-env in which the white car denotes the ego vehicle and blue cars are npc vehicles. Note that such a modification is only used for reward generation when using image and video based observations. Meanwhile, we also customize our opposite linguistic goals to adapt to the image and video based observations. To be specific, we define the opposite linguistic goal for image  and video based observation as ``\textit{White car collides with a blue car.}''. In this work, we don't fine-tune any large models to adapt to the Highway-env.

\subsection{Comparison with Baseline Methods}
\label{sec:baseline_compare}
We compare $3$ variants of our method (i.e. observation represented by image, video and text) with following baselines.
\begin{itemize}
    \item \textsc{GRAD} \cite{xi2022graph}. The latest state-of-the-art reinforcement learning method for the Highway-env\cite{highway-env}. It learns driving policy from a graph-based state representation using PPO\cite{schulman2017proximal} algorithm. The reward is a sum of 1) a constant surviving reward which is $0.2$; 2) a speed reward linearly mapped from the speed of $(20, 40)$ to $(0, 0.8)$.
    \item \textsc{Const.} Similar to \textsc{GRAD} except the reward only consists of a constant surviving reward in order to motivate the ego vehicle to survive as long as possible. 
    \item \textsc{DiLu} \cite{wen2023dilu}. The latest state-of-the-art large language model-based method. It leverages large language models to perform step-wise decision-making for autonomous driving task. 
\end{itemize}

\begin{table*}
\caption{Performance comparisons of all methods in different traffic situations of highway environments. All methods are only optimized on \texttt{lane-4-density-2} setting and evaluated on the \texttt{lane-4-density-2}, \texttt{lane-5-density-2.5}, and \texttt{lane-5-density-3}. The best results are highlighted in bold and the second-best results are marked with an underline. (SR: Success Rate, TD: Traveled Distance, RE: Rewards)}
\label{tbl:performance}
\begin{center}
\begin{tabular}{lccccccccccc}
\specialrule{0.12em}{0pt}{2pt}
& \multicolumn{3}{c}{lane-4-density-2}&& \multicolumn{3}{c}{lane-5-density-2.5}&& \multicolumn{3}{c}{lane-5-density-3}\\
\cline{2-4}\cline{6-8}\cline{10-12}
\specialrule{0em}{1pt}{1pt}
Method &SR$\uparrow$ &TD$\uparrow$ &RE$\uparrow$& &SR$\uparrow$ &TD$\uparrow$ &RE$\uparrow$& &SR$\uparrow$ &TD$\uparrow$ &RE$\uparrow$ \\
\specialrule{0.12em}{1pt}{1.5pt}
\textsc{GRAD}\cite{xi2022graph} &\underline{94.12} &\textbf{930.14} &\textbf{20.16}& &\underline{88.24} &\textbf{930.99} &\textbf{20.35} & &58.82 &\textbf{664.55} &\textbf{13.61}   \\
\textsc{Const.} &\textbf{100.0} &610.17 &6.39 & &\underline{88.24} &568.03 &5.80 & &52.94 &424.07 &4.18  \\
\textsc{DiLu}\cite{wen2023dilu} &70.00 &- &- & &65.00 &- &- & &35.00 &- &- \\
\specialrule{0.01em}{1pt}{1pt}
\textbf{LORD} image &\textbf{100.0} &\underline{694.20} &\underline{9.75} & &\underline{88.24} &\underline{652.02} &\underline{9.12} & &\underline{64.71} &578.41 &\underline{8.39} \\
\textbf{LORD} video &\textbf{100.0} &630.37 &7.24 & & \underline{88.24} &612.77 &7.67 & &\textbf{82.35} &\underline{599.86} &7.78 \\
\textbf{LORD} text &\textbf{100.0}  &682.24 &9.27 & &\textbf{94.12}  &630.53 &7.94 & &58.82 &493.24 &5.59\\
\specialrule{0.12em}{1.5pt}{0pt} 
\end{tabular}
\end{center}
\end{table*}

We optimize all methods on \texttt{lane-4-density-2} setting in Highway-env and evaluate them on various traffic situations, namely \texttt{lane-4-density-2}, \texttt{lane-5-\\density-2.5} and \texttt{lane-5-density-3}. Each evaluation is repeated $17$ times with different random seeds specified in the code repository\footnote{https://github.com/PJLab-ADG/DiLu} of DiLu~\cite{wen2023dilu}. We report Success Rate (SR), Traveled Distance (TD) and Rewards (RE) achieved by these methods in Table \ref{tbl:performance}. Success Rate (SR) is defined in \cite{wen2023dilu} where a success denotes that the ego vehicle survives over $30$ time steps without any collisions. Traveled Distance (TD) means how far the ego vehicle drives along the $x$ axis before a collision happens. We also adopt the reward function defined in \cite{xi2022graph} to calculate the Rewards (RE) as an additional metric to evaluate how well the ego vehicle drives.

\begin{wraptable}{r}{0.45\textwidth}
\caption{The performance of our LORD with addition of the speed reward using text based observation.}
\label{tbl:speed_exp}
\small
\begin{center}
\begin{tabular}{ccccccc}
\specialrule{0.12em}{0pt}{2pt}
\multicolumn{3}{c}{lane-4-density-2}& & \multicolumn{3}{c}{lane-5-density-3}\\
\cline{1-3}\cline{5-7}
\specialrule{0em}{1pt}{1pt}
SR$\uparrow$ &TD$\uparrow$ &RE$\uparrow$& &SR$\uparrow$ &TD$\uparrow$ &RE$\uparrow$ \\
\specialrule{0.12em}{1pt}{1.5pt}
94.12 &925.94 &19.57 &&64.71 &716.47 &14.32\\
\specialrule{0.12em}{0pt}{2pt}
\end{tabular}
\end{center}
\end{wraptable} 

As shown in Table~\ref{tbl:performance}, in the in-domain training environment \texttt{lane-4-density-2}, all variants of our method achieve $100\%$ success rate (SR), which is $5.88\%$ higher than the RL-only baseline \textsc{GRAD} and $30\%$ higher than the LLMs-only baseline \textsc{DiLu}. LORD also shows strong generalization performance in unseen out-of domain scenarios. In particular, for \texttt{lane-5-density-2.5}, our method with image and video based observation achieves $88.24\%$ SR, which is same as \textsc{GRAD} and \textsc{Const.}, and $23.24\%$ higher than \textsc{DiLu}. With text based observation, our method outperforms all the counterpart methods by achieving further $5.88\%$ SR improvement. In the most complex traffic situation, i.e. \texttt{lane-5-density-3}, our method with image and video based observation outperforms the best performing counterpart method (GRAD) in terms of SR by $5.89\%$ and $23.53\%$, respectively. These comparisons demonstrate that the driving policy learned by our LORD not only helps the ego vehicle to better avoid collisions in seen environments but also generalizes well to unseen environments.

For the metrics of Traveled Distance (TD) and Rewards (RE), our method also outperforms the \textsc{Const.} baseline in all traffic situations, showing that our large model based reward motivates the ego vehicle to drive faster than a constant survival reward. We note that \textsc{GRAD} achieves much higher TD and RE. We hypothesize it is because \textsc{GRAD} directly optimizes RE in which the speed reward encourages the ego vehicle to speed up. To validate our hypothesis, we conduct an experiment by adding the additional speed reward into our LORD method (observation represented by text). Since the speed is a concrete information and can be easily obtained from the sensory inputs, adding the speed reward is straightforward. We report the results in Table~\ref{tbl:speed_exp} and the results show that our method achieves $51.92$ higher TD and $0.71$ higher RE than the \textsc{GRAD} baseline in the challenging \texttt{lane-5-density-3} setting.

\subsection{Deep Dive into LORD}
\label{sec:reward_analyze}
\begin{figure}[ht!]
\centering
    \begin{subfigure}{0.48\textwidth}
        \includegraphics[width=\textwidth]{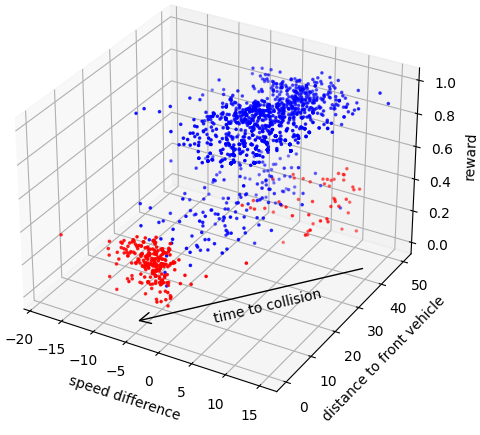}
        \caption{LORD (image)}
        \label{fig:r_i}
    \end{subfigure}
    \begin{subfigure}{0.48\textwidth}
        \includegraphics[width=\textwidth]{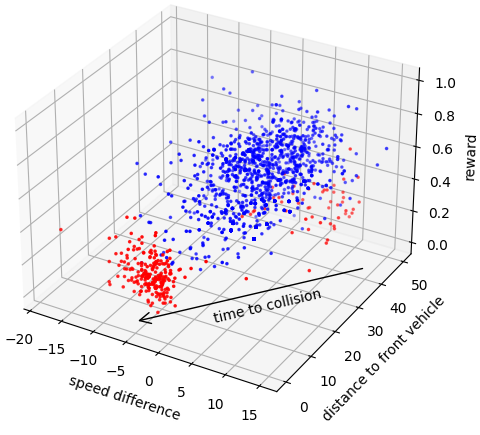}
        \caption{LORD (video)}
        \label{fig:r_v}
    \end{subfigure}
    \\
    \begin{subfigure}{0.48\textwidth}
        \includegraphics[width=\textwidth]{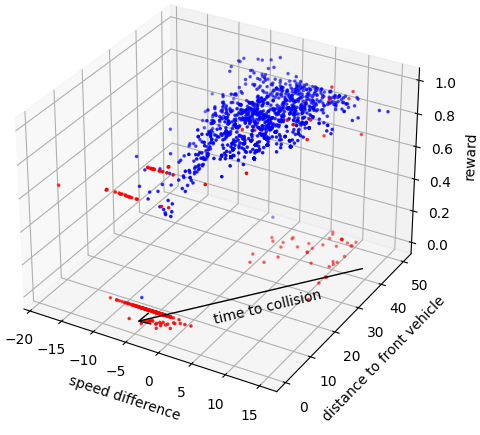}
        \caption{LORD (text)}
        \label{fig:r_t}
    \end{subfigure}
    \begin{subfigure}{0.48\textwidth}
        \includegraphics[width=\textwidth]{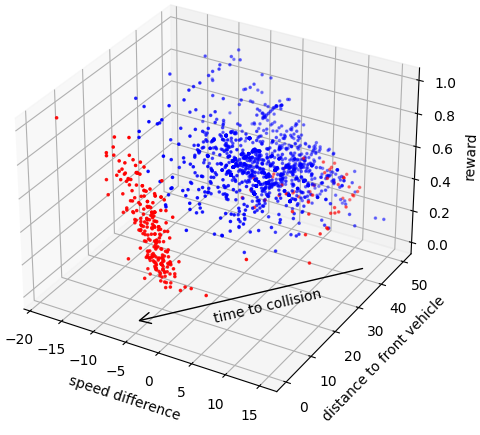}
        \caption{GRAD\cite{xi2022graph}}
        \label{fig:r_g}
    \end{subfigure}
\caption{Illustrations of the reward values generated by our LORD under various observation representations and GRAD\cite{xi2022graph} for different states. We distinguish different states in terms of the ego vehicle's distance to its nearest front vehicle and their speed difference. In this way, time to collision can be roughly estimated. Blue points denote collision-free states while red points indicate the ego vehicle collides with other vehicles. }
\label{fig:rewards}
\end{figure}

Table~\ref{tbl:performance} shows comparison of the $3$ variants of our method as detailed in Sec.~\ref{sec:approach}. Overall, our method with different observation representations achieve similar performance. Notably, LORD achieves $100\%$ success rate in the training environment \texttt{lane-4-density-2} under all image, video and text based observations, indicating the effectiveness of our LORD method. In terms of generalization ability, LORD with text based observation achieves higher success rate in \texttt{lane-5-density-2.5} while image and video based observation helps LORD perform better in \texttt{lane-5-density-3} environment.

To have an in-depth understanding of the remarkable performance achieved by LORD, we illustrate in Fig.~\ref{fig:rewards} the reward values generated by LORD for different states. Since the state of the ego vehicle is a high dimensional vector that includes position, speed and orientation information of all vehicles in the scenario, it is impractical to visualize the rewards with respect to the states directly. It is also extremely difficult to evaluate how good a state is and how well the reward values align with the goodness of the states. Therefore, we instead choose $2$ features to represent a state, namely the distance to the nearest front vehicle and the vehicle's relative speed comparing to the ego vehicle. In this way, we can estimate their time to collision (ttc), and  use the ttc as a surrogate metric to measure how good the state is. In addition, we also distinguish the collision and non-collision states by coloring them in red and blue, respectively.

As shown in Fig.~\ref{fig:rewards}, the collision states denoted by the red points are centered as expected at the area having small distance to the front vehicle and negative speed difference, which means that the ego vehicle collides with the front vehicle at a higher speed. From Fig.~\ref{fig:r_i}, \ref{fig:r_v} and \ref{fig:r_t}, we can see that our LORD consistently assigns smaller rewards to these collision states. For non-collision states denoted by the blue points, we also observe a decrease in reward values along the trend over the time to collision, especially when the distance is small and the speed difference is negative as such a collision is more likely to happen. We note that when the ego vehicle is far away from the front vehicle or drives slower than the front vehicle, more tailored features might be needed to evaluate the state in such cases.
Collisions could also happen as the red points in the area of large distance and positive speed difference show. In comparison, GRAD \cite{xi2022graph} gives high rewards even to the collision states as Fig.~\ref{fig:r_g} shows. The rewards also show a clear upward trend when the speed difference becomes negative, i.e., the ego vehicle drives faster than its front vehicle. It is expected as GRAD adopts a speed reward but such a reward strategy doesn't encourage a safe driving policy.

\subsection{Effectiveness of Opposite Reward Design}
\label{sec:reward_ablation}
To validate our opposite reward design, we conduct ablation study to compare the $3$ variants of our method with the ones using the corresponding target goals. To be specific, when using image and video to represent the observation, we set the target goal as ``\textit{White car drives safely.}''.  For the setting using text to represent the observation, the target goal is set as ``\textit{Ego is driving safely.}''. In addition, when using the target goal, we define the reward $r_t$ at time step $t$ as the cosine similarity between the state $o_t$ and the target goal, i.e. $r_t = similarity(o_t, goal)$ where the $similarity(\cdot, \cdot)$ function is defined in Eq.~\ref{eq:sim}.

\begin{table*}
\caption{An ablation study of our method using opposite goals versus target goals. Our approach witnesses the opposite goals show significantly improved generalization performance on challenging scenarios. }
\label{tbl:ablation_goal}
\begin{center}
\begin{tabular}{llccccccccccc}
\specialrule{0.12em}{0pt}{2pt}
&& \multicolumn{3}{c}{lane-4-density-2}&& \multicolumn{3}{c}{lane-5-density-2.5}&& \multicolumn{3}{c}{lane-5-density-3}\\
\cline{3-5}\cline{7-9}\cline{11-13}
\specialrule{0em}{1pt}{1pt}
State &Goal &SR$\uparrow$ &TD$\uparrow$ &RE$\uparrow$& &SR$\uparrow$ &TD$\uparrow$ &RE$\uparrow$& &SR$\uparrow$ &TD$\uparrow$ &RE$\uparrow$ \\
\specialrule{0.12em}{1pt}{1.5pt}
\multirow{2}{*}{Image} &oppo. &\textbf{100.0} &\textbf{694.20} &9.75 & &\textbf{88.24} &\textbf{652.02} &9.12 & &\textbf{64.71} &\textbf{578.41} &\textbf{8.39}\\
 &target &70.59 &672.92 &\textbf{10.03} & &47.06 &639.69 &\textbf{10.01} & &29.41 &527.48 &7.78 \\
\specialrule{0.01em}{1pt}{1pt}
\multirow{2}{*}{Video} &oppo. &\textbf{100.0} &630.37 &7.24 & & \textbf{88.24} &612.77 &7.67 & &\textbf{82.35} &\textbf{599.86} &\textbf{7.78} \\
 &target &\textbf{100.0} &\textbf{679.10} &\textbf{9.14} & &76.47 &\textbf{646.01} &\textbf{8.34} & &58.82 &519.54 &6.32 \\
\specialrule{0.01em}{1pt}{1pt}
\multirow{2}{*}{Text} &oppo. &\textbf{100.0}  &\textbf{682.24} &\textbf{9.27} & &\textbf{94.12}  &\textbf{630.53} &\textbf{7.94} & &\textbf{58.82} &\textbf{493.24} &\textbf{5.59} \\
 &target &\textbf{100.0} &672.88 &8.87 & &76.47 &583.62 &7.52 & &23.53 &305.66 &3.81 \\
\specialrule{0.12em}{1.5pt}{0pt} 
\end{tabular}
\end{center}
\end{table*}

The comparisons of the methods using our opposite goals and target goals are shown in Table~\ref{tbl:ablation_goal}. Overall, in terms of success rate (SR), all variants of our method with opposite reward design consistently outperform their counterparts using target goals in all traffic situations, indicating that our opposite reward design is more effective in learning a safe driving policy. In addition, while the method using target goals achieves competitive performance in in-domain traffic situation \texttt{lane-4-density-2}, its performance degrades a lot in more difficult out-of-domain traffic situations. In particular, with video and text based observation, the method using target goals achieve $100\%$ SR in \texttt{lane-4-density-2}. However, its SR drops to $76.47\%$ in \texttt{lane-5-density-2.5} and finally to $58.82\%$ and $23.53\%$ respectively in \texttt{lane-5-density-3}. It demonstrates that the target goal reward design is not generalizable for autonomous driving task. Comparing to our method, in \texttt{lane-5-density-2.5}, we note that the method with target goal reward design has a slightly higher rewards (RE) under image and video based observation. However, its success rates (SR) are much lower. Specifically, while the target goal reward design obtains $0.89$ and $0.67$ higher RE under image and video observation, the SR is $41.18\%$ and $11.77\%$ lower than ours. We explain this as the method using target goals learns to accelerate but ignores the importance of avoiding collisions. More importantly, in the most difficult traffic situation \texttt{lane-5-density-3}, our opposite reward design outperforms the target goal reward design in all metrics, showing the superiority of our method in generalization.

\subsection{Qualitative Results}
\label{sec:qual_res}
\begin{figure}[ht!]
    \centering
    \includegraphics[width=1\textwidth]{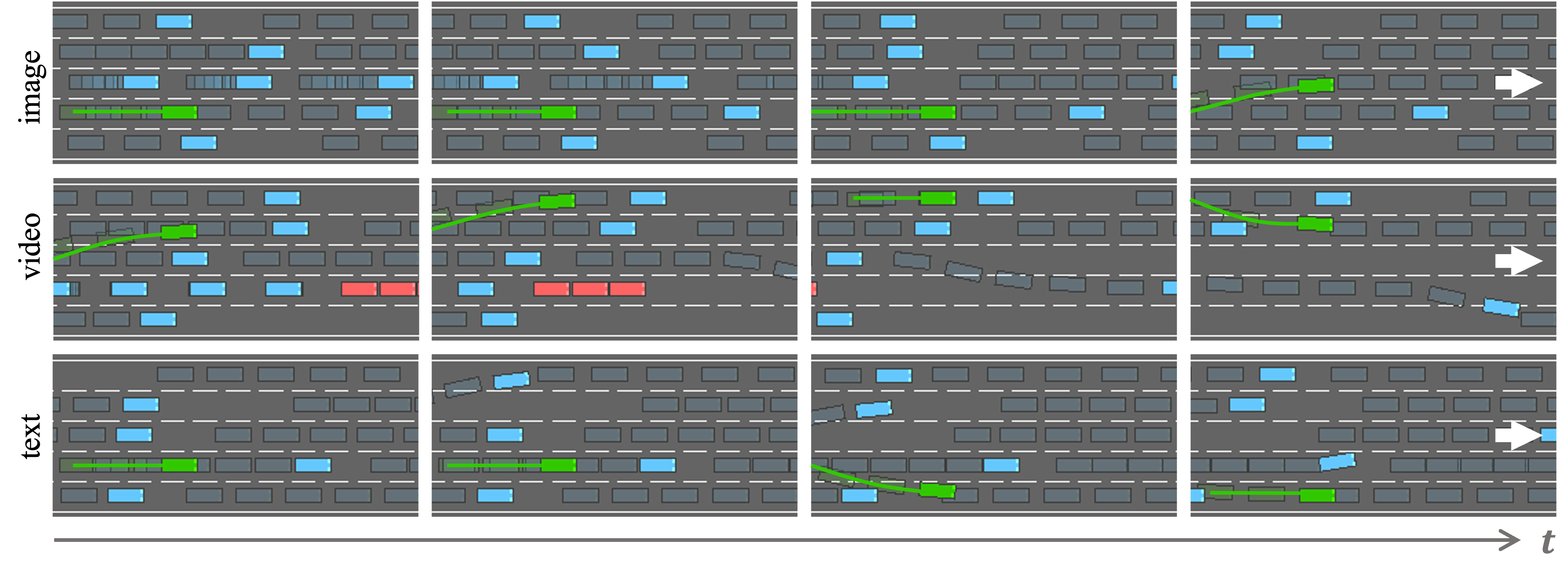}
    \caption{Illustrations of how the driving policy learned by our LORD with image, video and text based observation performs in the \texttt{lane-5-density-3} setting of Highway-env. The ego vehicle is colored in green with a line depicting its past trajectory. The ego vehicle behaves properly in the congested traffic scenarios. }
    \label{fig:vis_res}
\end{figure}

Fig.~\ref{fig:vis_res} depicts how the driving policy learned by our LORD in  \texttt{lane-4-density-2} setting of Highway-env performs in more challenging \texttt{lane-5-density-3} environments with image, video and text based observation respectively. As shown in the figure, the ego vehicle learns to behave properly in the congested traffic scenarios even if it hasn't encountered such situations before. For example, from the top row of Fig.~\ref{fig:vis_res}, we can see that the ego vehicle chooses to follow the front vehicle when there is no room for a lane change. Once the ego vehicle surpasses all left-lane vehicles, it makes a left lane change to gain more space. The ego vehicle also learns to overtake its front vehicle with video based observation as illustrated in the middle row. Similarly, with text based observation, the ego vehicle also succeeds in improving its situation by changing its lane to the right. We provide more qualitative results in the supplementary materials.

%% file: eccv2024/Sections/conclusion.tex
\section{Conclusion and Future Work}
In this paper, we introduce a novel large models based opposite reward design (LORD) framework for autonomous driving. LORD presents opposite reward design through undesired linguistic goals for efficient use of large pretrained models as zero-shot reward mechanism since such undesired goals are more tractable and comprehensible for large pretrained models compared to desired ones.
We leverage large pretrained image, video and language models with a cosine distance objective in our reward function. We integrate LORD with reinforcement learning algorithms to perform autonomous driving tasks. 
Extensive experiments on the closed-loop autonomous driving tasks show the efficacy of the opposite reward design mechanism over the desired target goal reward design. Moreover, LORD achieves improved generalization performance over the counterpart reinforcement learning  and language model based  methods. 

Our experiments are currently confined to Highway-env simulation as baseline approaches only report their performance in this simulation. We also note that the exceptional abilities of large pretrained models have not been fully utilized due to the simulated images and videos we input and thereby limits the performance of our approach. We will assess our approach on more environments in our future work. In this work, our study has evolved around the ``\textit{collision}'' as our opposite linguistic goal.
In reality, there are numerous undesired driving behaviors and situations the autonomous agents should avoid, such as running a red light, occupying an emergency lane or violating other traffic rules. With our opposite reward design, we can add these undesired behaviors as additional opposite linguistic goals into our LORD framework so that we can further optimize the driving policy. In our future work, we will explore these in more sophisticated environments as Highway-env does not contain such detailed information.

%% file: eccv2024/Sections/appendix.tex
\section*{Appendix}
\appendix
\section{Implementation Details}
\subsection{Detailed Setup of Highway-env}
We follow the code repository \footnote{https://github.com/zerongxi/graph-sdc} of GRAD baseline to setup Highway-env. Table~\ref{tbl:highway-setup} shows our customized configurations and we use default values for other parameters. During training, we set \texttt{lane\_count} as $4$ and \texttt{vehicles\_density} as $2$ to train all methods in \texttt{lane-4-density-2} setting. In addition, we set \texttt{duration} as $60$ to train the agent to address long-horizon tasks. During testing, we set \texttt{lane\_count} and \texttt{vehicles\_density} accordingly and change \texttt{duration} to $30$ to evaluate the success rate of all methods in \texttt{lane-4-density-2}, \texttt{lane-5-density-\\2.5} and \texttt{lane-5-density-3} settings.
\begin{table} 
\centering
\caption{Configurations of Highway-env for training.}
\label{tbl:highway-setup}
\begin{tabular}{ll}
\specialrule{0.12em}{0pt}{2pt}
Parameter                       & Value    \\
\specialrule{0.1em}{1pt}{1pt}
observation                     &          \\
\ --type                        & Kinematics \\
\ --features                    & [presence, x, y, vx, vy, cos\_h, sin\_h, heading]\\
\ --absolute                    & True \\
\ --normalize                   & True \\
\ --vehicles\_count             & 33 \\
\ --see\_behind                 & True \\
action                          &  \\
\ --type                        & DiscreteMetaAction \\
\ --target\_speeds              & [20, 25, 30, 35, 40] \\
duration                        & 60 \\
ego\_spacing                    & 4  \\
lane\_count                     & 4  \\
vehicles\_density               & 2  \\
\specialrule{0.12em}{0pt}{2pt}
\end{tabular}
\end{table} 

\subsection{Observation Design}
To enable a more efficient use of large pretrained models as zero-shot reward models, we empirically adopt the following observation designs as inputs to the large pretrained models. (1) For image based observation, we adopt the simulated image rendered by Highway-env with the parameter \texttt{scaling} being set to $10$. We then replace the rectangles used to represent the ego and npc vehicles with more photorealistic car images. We further remove the image background and we crop the image to the size of $224 \times 224$ centered on the ego vehicle. (2) For video based observation, we stack the latest $30$ consecutive image based observations with a $15$Hz frequency. (3) For text based observation, we only pay attention to the nearby vehicles that are within $5 \times ego\_speed$ meters of the ego vehicle and drive on the same, left or right lane of the ego vehicle. We then calculate the ego vehicle's time to collision (ttc) to each of these attended vehicles. If a vehicle drives on the same lane of the ego vehicle and the ttc is smaller than $5$s, we describe it in our text based observation by ``\textit{A collision will be happening in \{ttc\}s.}''. Otherwise, we give a description of ``\textit{No foreseeable collision in $5$s.}''. We also describe conditional collisions by ``\textit{A collision would happen in \{ttc\}s if ego makes a left/right lane change.}''.  Examples of the three types of observations can be found in Fig.~\ref{fig:safe_obs} and Fig.~\ref{fig:collision_obs}.

\section{Case Study}
\subsection{Rewards from Different Observations}

\begin{figure}
\centering
\includegraphics[width=0.8\textwidth]{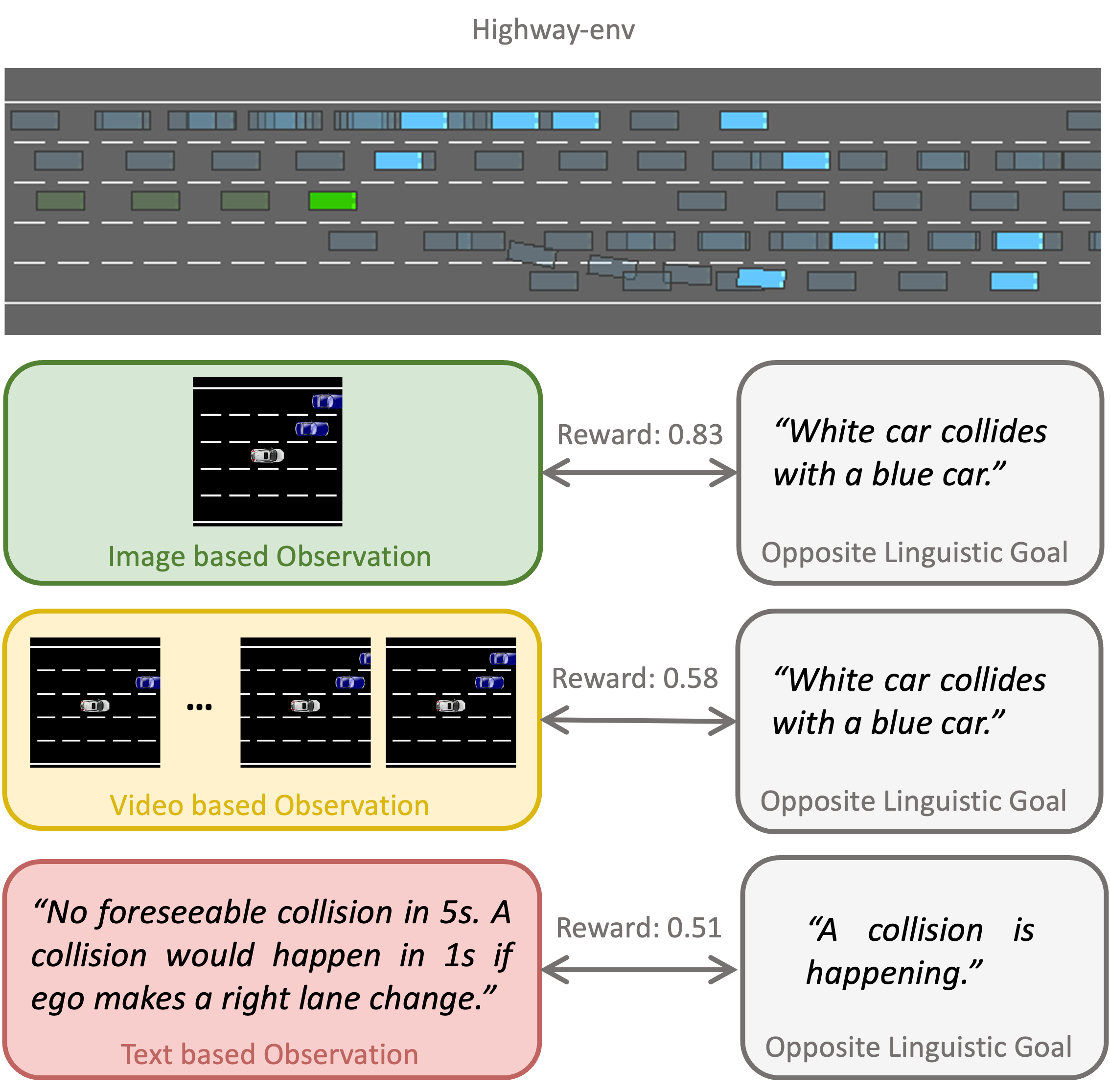}
\caption{An example of our image, video and text based observations and the corresponding rewards for a non-collision state.}
\label{fig:safe_obs}
\end{figure}

\begin{figure}
\centering
\includegraphics[width=0.8\textwidth]{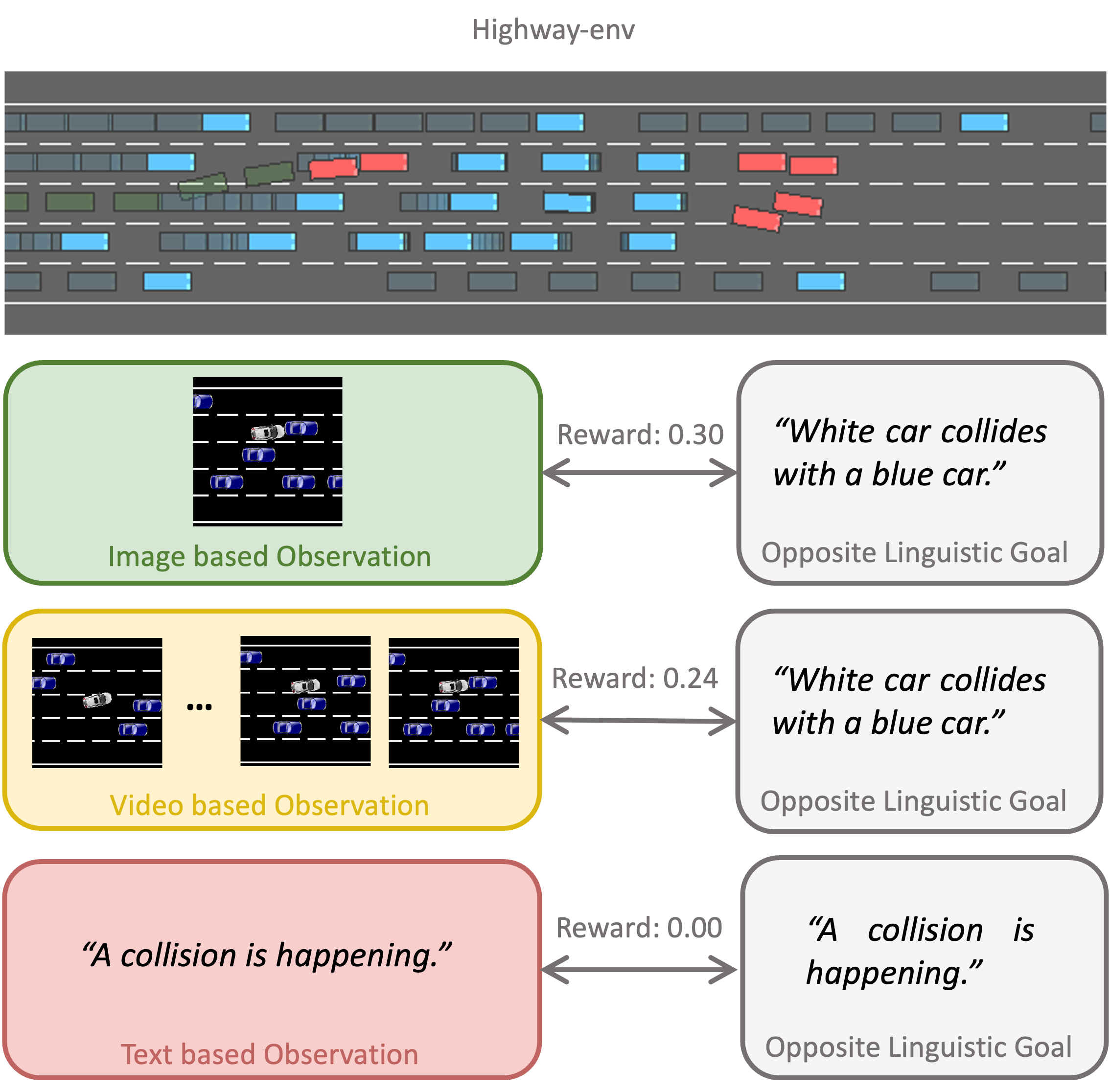}
\caption{An example of our image, video and text based observations and the corresponding rewards for a collision state.}
\label{fig:collision_obs}
\end{figure}

Fig.~\ref{fig:safe_obs} and Fig.~\ref{fig:collision_obs} present the rewards we get from different observations for a non-collision and a collision state respectively. While the reward values are nonidentical across different observations, they are all higher for the non-collision state compared to the collision one. In this way, the ego vehicle can distinguish the dangerous states and learn a safe driving policy.

\subsection{Qualitative Results}
\begin{figure}[ht]
\centering
\begin{tabular}{c}
\includegraphics[width=0.97\textwidth]{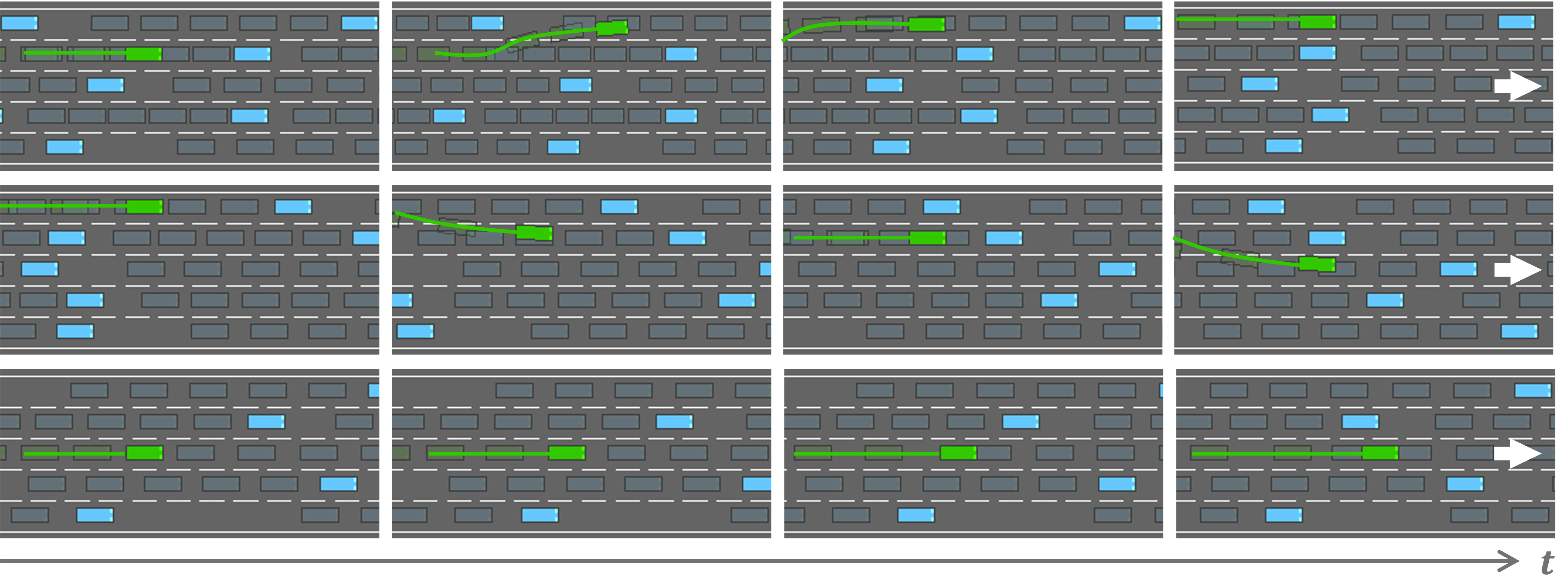}\\
(a) LORD with image based observation. \\\\
\includegraphics[width=0.97\textwidth]{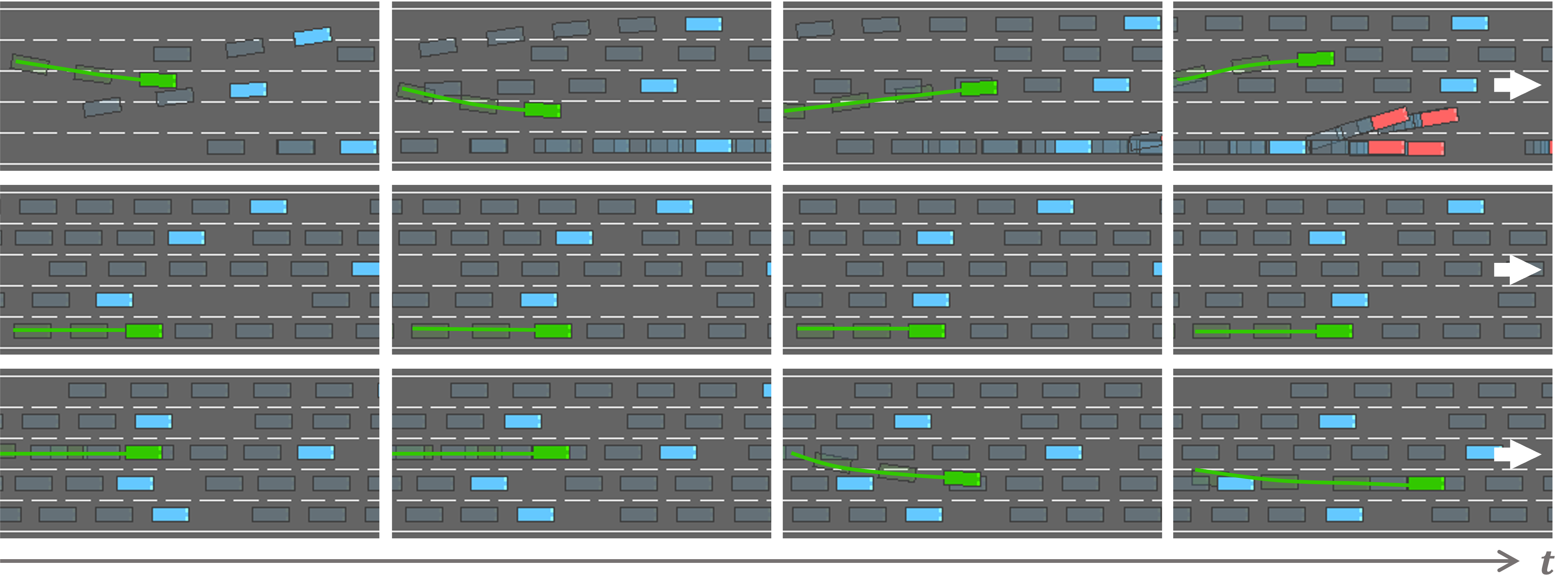}\\
(b) LORD with video based observation. \\\\
\includegraphics[width=0.97\textwidth]{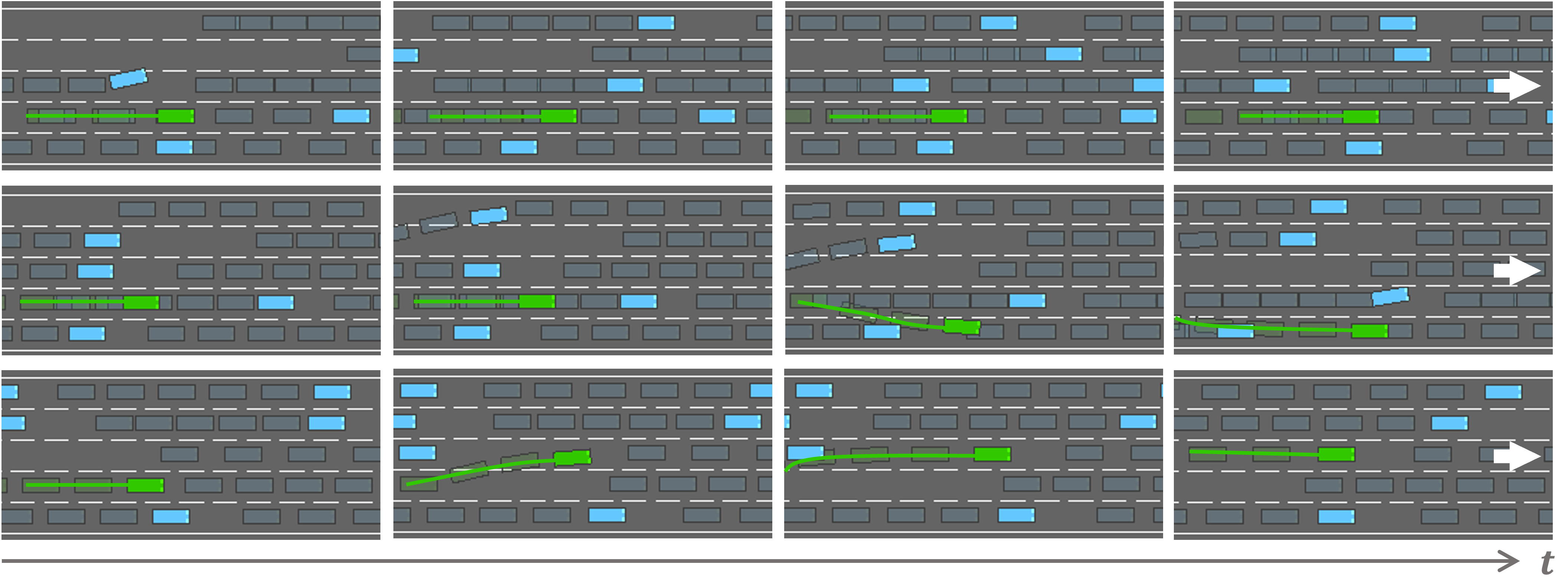}\\
(c) LORD with text based observation. \\\\
\end{tabular}
\caption{The driving policy learned by our LORD with image, video and text based observation.}
\label{fig:lord_vis}
\end{figure}

Fig.~\ref{fig:lord_vis} shows more examples of how the driving policy learned by our LORD with image, video and text based observation performs in the \texttt{lane-5-density-3} setting of Highway-env. We can observe that the ego vehicle learns diverse ways to avoid collisions in congested traffic scenarios. The results shall be better viewed in the supplementary videos.

